# A Divergence Bound for Hybrids of MCMC and Variational Inference and an Application to Langevin Dynamics and SGVI

**Justin Domke** [1]

## Abstract

Two popular classes of methods for approximate inference are Markov chain Monte Carlo (MCMC) and variational inference. MCMC tends to be accurate if run for a long enough time, while variational inference tends to give better approximations at shorter time horizons. However, the amount of time needed for MCMC to exceed the performance of variational methods can be quite high, motivating more fine-grained tradeoffs. This paper derives a distribution over variational parameters, designed to minimize a bound on the divergence between the resulting marginal distribution and the target, and gives an example of how to sample from this distribution in a way that interpolates between the behavior of existing methods based on Langevin dynamics and stochastic gradient variational inference (SGVI).

## 1. Introduction

One of the simplest Markov chain Monte Carlo (MCMC) methods is Langevin dynamics (Grenander & Miller, 1994; Robert & Casella, 2004, Sec. 7.8.5) where one repeats gradient steps with injected Gaussian noise. Namely, one iterates

$$z \leftarrow z + \frac{\epsilon}{2} \nabla_z \log p(z) + \sqrt{\epsilon} \eta, \qquad (1)$$

where $\eta$ is sampled from a standard Gaussian distribution and $\epsilon$ is a step-size that may decay over time. To get exact samples, a Metropolis-Hastings rejection step should be used. However, as the step-size $\epsilon$ becomes smaller, the acceptance probability goes to one, and so this can be disregarded. In the Bayesian inference setting, $z$ denotes unknown parameters and $p(z)$ represents a posterior over them. Computing $p(z)$ thus requires a full pass over the dataset. The idea of Stochastic Gradient Langevin Dynamics (Welling & Teh, 2011) is to get an unbiased estimate of the gradient of $\log p(z)$ by approximating the posterior using a minibatch. Since the gradient term is multiplied by $\epsilon$, the variance of its error has order $\epsilon^2$, while the variance of $\sqrt{\epsilon} \eta$ has order $\epsilon$. Thus, with a small step $\epsilon$, the injected noise dominates, meaning one can might expect these to have a stationary distribution close to the target $p(z)$.

[1]College of Computing and Information Sciences, University of Massachusetts, Amherst, USA. Correspondence to: Justin Domke <domke@cs.umass.edu>.

The idea of variational inference is to posit a simple family of distributions $q_w(z)$ and then optimize $w$ to minimize the KL-divergence from $q_w(z)$ to the target $p(z)$. (In a Bayesian setting, this is equivalent to maximizing the evidence lower bound.) If one were to use plain gradient ascent to perform this optimization, this leads to the iterate of

$$w \leftarrow w + \frac{\epsilon}{2} \nabla_w \mathbb{E}_{q_w(Z)}[\log p(Z) - \log q_w(Z)], \qquad (2)$$

where the gradient step $\epsilon$ is divided by two for convenience, and may again decay over time. In practice, doing gradient ascent like this has two difficulties. First, in the Bayesian inference setting, computing $\log p(z)$ requires a full pass over the data. This can be approximated without bias using a minibatch. The second difficulty stems from the *expectation* of $\log p(z)$ with respect to $q_w(z)$ in Eq. 2. The gradient of this expectation can be approximated in different ways using samples from $q_w(Z)$ (Section 3.2). Since this gives unbiased estimates of the gradient, stochastic optimization methods, such as stochastic gradient descent, will converge to a local optima when the step-size becomes small.

Intuitively, both MCMC and variational methods can be thought of as trying to find a high-probability region of $\log p(z)$, with different strategies to encourage entropy to get coverage of the distribution. In MCMC, entropy is created by randomness in the Markov chain, while in variational methods the KL-divergence directly measures the entropy of the variational distribution. It is natural to think that hybrid methods might employ fractions of these two strategies. This paper does so by defining a *distribution over the parameters* $w$ of the approximating family $q(z|w)$. This distribution $q(w)$ is designed to minimize a bound on the KL-divergence of the resulting marginal distribution $q(z)$ to the target $p(z)$. Intuitively, when defining $q(w)$, there are two ways to encourage entropy, namely to give higher density to vectors $w$ where $q(Z|w)$ has high entropy or to encourage entropy over $w$ itself.

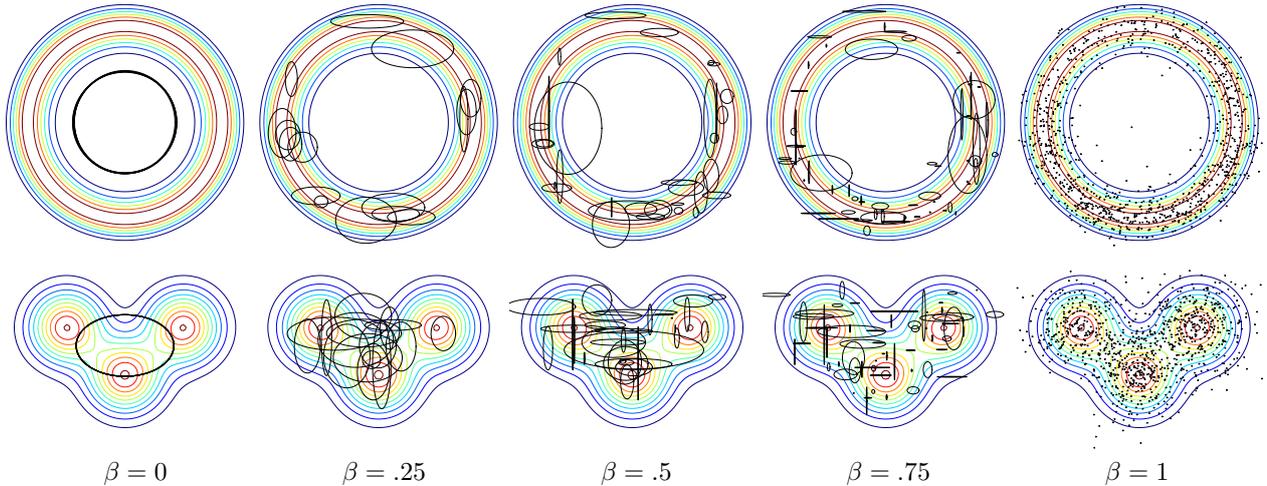

*Figure 1.* Inference on toy distributions. Colored contours show the target density $p(z)$. Sampled weights $w$ are drawn equally from $5 \times 10^5$ iterations and pictured as ellipsoids at one standard deviation. Fewer $w$ are shown for smaller $\beta$ to avoid visual clutter.

Given a distribution $q(w, z)$ and a target distribution $p(z)$, this paper uses two bounds on the marginal divergence $D_{\text{true}} = KL\left(q(Z)\|p(Z)\right)$. First, one can upper-bound by the conditional divergence $D_0 = KL\left(q(Z|W)\|p(Z)\right)$. Second, one can "adjoin" some distribution $p(w|z)$ to $p(z)$, and bound $D_{\text{true}}$ by the joint divergence $D_1 = KL\left(q(W, Z)\|p(W, Z)\right)$. This paper takes a convex combination of these two bounds, i.e. uses the family of bounds $D_\beta = (1 - \beta)D_0 + \beta D_1$ . For any $\beta$ between 0 and 1, the distribution $q(w)$ is derived that minimizes this bound for fixed $p$ and a fixed approximating family $q(z|w)$ (Theorem 3). Then, one can sample from this distribution over $w$ by (stochastic) Langevin dynamics. This turns out to lead to the iterate

$$w \leftarrow w + \frac{\epsilon}{2}\nabla_w\Big(\mathbb{E}_{q_w}[\log p(Z) + (\beta - 1)\log q_w(Z)] + \beta \log r(w)\Big) + \sqrt{\epsilon}\beta\eta, \quad (3)$$

where $r(w)$ is related to the adjoining distribution $p(w|z)$ and may be thought of as a "base measure" [1] over the space of $w$. Of course, exactly computing this gradient is intractable in general, but the same strategies used for variational inference can be used to efficiently compute an un-biased estimate.

It can be shown that as $\beta \to 0$, the solution concentrates in the optima of the variational optimization, and thus the sampling algorithm reduces to gradient ascent on the variational objective. Meanwhile, as will be discussed in Sections 2.3 and 5 it is best to make $r$ a function of $\beta$ such that $w$ with high entropy have less density when $\beta$ approaches 1. When this is true, then as $\beta \to 1$, only $w$ indexing highly concentrated distributions over $z$ retain density, and so sampling from $q(w)$ is essentially just a reparameterization of sampling from $p(z)$, and the algorithm reduces to Langevin dynamics.

Thus, for the extreme cases of $\beta = 0$ and $\beta = 1$, this algorithm reduces to variational inference and Langevin dynamics, respectively. Informally, at these extremes, the algorithm is "fast but approximate" and "slow but accurate". In the intermediate range, the algorithm exhibits new behavior with a fine-grained trade-off between speed and accuracy. Thus, this approach gives a smooth continuum of algorithms with different priorities of speed and accuracy.

### 1.1. Notation

This paper uses three notational conventions that are not universal. First, the conditional probability $q(z|w)$ will alternatively be written as $q_w(z)$ when convenient. Second, $A \simeq B$ indicates that $A$ and $B$ have the same expected value or (if $A$ is a constant) that $B$ is an unbiased estimator of $A$. Finally, upper- and lower-cases indicate what terms are random variables in a KL-divergence. So, $KL\left(q(Z|w)\|p(Z|w)\right) = \mathbb{E}_{q(Z|w)}\log\left(q(Z|w)/p(Z|w)\right)$ is the divergence between $q(z|w)$ and $p(z|w)$ for a fixed $w$, while $KL\left(q(Z|W)\|p(Z|W)\right) =$

---

[1]To see why $r$ is necessary, consider a reparameterization of $w$. In order for the marginal distribution $q(z)$ to stay the same, the density of the base measure must be transformed corresponding to the reparameterization. Put another way, without $r(w)$, the results of inference would depend on the parameterization of $w$, even with an arbitrarily small step-size.

$\mathbb{E}_{q(W,Z)} \log\left(q(Z|W)/p(Z|W)\right)$ is a standard conditional divergence.

# 2. Information Theoretic Results

This section derives a few results from an information theoretic viewpoint, without any particular regard for form of the target distribution, or how one might sample from it. Proofs are given in the appendix.

## 2.1. Preliminaries

Let $p(z)$ be the target distribution. For simplicity, $z$ is treated here as a continuous variable, although the results in this section remain true if it is discrete. There is some fixed set of conditional distributions $q(z|w)$, which may also be written as $q_w(z)$. In principle, one might like to know how to set $q(w)$ such that the resulting marginal distribution over $z$, is close to $p(z)$, as measured by the KL-divergence,

$$D_{\text{true}} = KL\left(q(Z)\|p(Z)\right) = \mathbb{E}_{q(Z)} \log \frac{q(Z)}{p(Z)}. \quad (4)$$

This quantity is difficult to control directly, since the marginal $q(z)$ typically cannot be evaluated in closed form. One bound comes from the conditional KL-divergence, as stated in the following Lemma.

**Lemma 1.** *The divergence from $q(z)$ to $p(z)$ is*

$$KL\left(q(Z)\|p(Z)\right) = \underbrace{KL\left(q(Z|W)\|p(Z)\right)}_{D_0} - I_q[W, Z], \quad (5)$$

*where* $D_0 = \mathbb{E}_{q(W,Z)} \log\left(q(Z|W)/p(Z)\right)$ *is the conditional divergence and* $I_q = KL\left(q(Z,W)\|q(Z)q(W)\right)$ *denotes mutual information under $q$.*

A second bound comes from adjoining some distribution $p(w|z)$ to $p(z)$. Then, the following lemma is essentially just a re-statement of the chain rule for the KL-divergence (Cover & Thomas, 2006, Thm. 2.5.3).

**Lemma 2.** *The KL-divergence between $q(z)$ and $p(z)$ can be written as*

$$KL(q(Z)\|p(Z)) = \underbrace{KL\left(q(W,Z)\|p(W,Z)\right)}_{D_1} - KL\left(q(W|Z)\|p(W|Z)\right). \quad (6)$$

Since the mutual information in Eq. 5 and the conditional divergence on the right of Eq. 6 are both non-negative, both $D_0$ and $D_1$ are upper-bounds on $D_{\text{true}}$. Note that if $p(w|z) = q(w)$, then $D_1 = D_0$, and so Lemma 1 follows from Lemma 2.

## 2.2. Divergence Bound and its Minimizer

This paper is based on convex combination of $D_0$ and $D_1$, parameterized by some $\beta \in [0,1]$, namely

$$D_\beta = (1-\beta)D_0 + \beta D_1. \quad (7)$$

Since $D_0$ and $D_1$ are upper-bounds, so is $D_\beta$. The following theorem gives the distribution $q(w)$ to minimize $D_\beta$.

**Theorem 3.** *For fixed values of $\beta$ and $p(w|z)$, the distribution $q(w)$ that minimizes $D_\beta$ is*

$$q^*(w) = \exp\left(s(w) - A\right) \quad (8)$$
$$A = \log \int_w \exp s(w)$$
$$s(w) = \log p(w) - KL\left(q(Z|w)\|p(Z|w)\right)$$
$$\quad - \left(\beta^{-1} - 1\right) KL\left(q(Z|w)\|p(Z)\right).$$

*Moreover, at $q^*$, the objective value is $D_\beta^* = -\beta A$.*

Since $-\beta A$ is an upper-bound on the KL-divergence, $A$ must be non-positive. This can also be seen directly, since it can be written as $A = \log \int_w p(w) \exp(-KL\left(q(Z|w)\|p(Z|w)\right) - \left(\beta^{-1} - 1\right) KL\left(q(Z|w)\|p(Z)\right))$, and the inner divergences as well as $\beta^{-1} - 1$ are non-negative.

To understand the connection of this bound to variational inference and MCMC, one can look at behavior where $\beta = 0$ or where $\beta = 1$. For the former case, one obtains a result closely related to the solution to the variational inference problem.

*Remark* 4. In the limit where $\beta \to 0$ the divergence bound at the optimal $q^*$ becomes

$$\lim_{\beta \to 0} D_\beta^* = \inf_w KL\left(q(Z|w)\|p(Z)\right). \quad (9)$$

Similarly, when $\beta \to 0$, $q^*(w)$ concentrates at the minimizer(s) of this divergence. (Formally, with multiple global optima with the same divergence, $q(w)$ will concentrate equally around all minimizers.)

For the case where $\beta = 1$, it is easy to see that

$$D_1^* = -\log \int_w p(w) \exp\left(-KL\left(q(Z|w)\|p(Z|w)\right)\right), \quad (10)$$

which will be zero if $p(w)$ is only supported on $w$ where the divergence between $q(z|w)$ and $p(z|w)$ is zero. This may initially seem like a strange condition, given that $p(w)$ results from both the target distribution $p(z)$ and the adjoined distribution $p(w|z)$. However, the next section gives more specifics.

## 2.3. Specific Form for $p(w|z)$

The results in the previous section were written without without considering the specific form of $p(w|z)$. This paper takes the approach of defining

$$p(w|z) = \frac{r(w)q(z|w)}{r_z}, \tag{11}$$

where $r(w)$ is a possibly improper distribution over $w$ and $r_z = \int_w r(w)q(z|w)$ is a normalizer. Note that $p(w|z)$ is still well-defined as long as the integral defining $r_z$ exists.

Here, we assume for convenience that $r_z$ is a constant, not depending on $z$. Enforcing $r_z$ to be constant essentially means choosing $r(w)$ in such a way that it doesn't "favor" any $z$ over any other since if $q(w) \propto r(w)$ then $q(z)$ is uniform over $z$.

**Lemma 5.** *If $p(w|z) = r(w)q(z|w)/r_z$ and $r_z$ is a constant, then the solution in Thm. 3 holds with*

$$\begin{aligned} s(w) = \log r(w) - \log r_z \\ + \mathbb{E}_{q_w(Z)}[\beta^{-1} \log p(Z) + (1 - \beta^{-1}) \log q_w(Z)]. \end{aligned} \tag{12}$$

This gives a distribution over $q(w)$ that can be written in various equivalent ways, such as

$$\begin{aligned} q^*(w) &\propto r(w) \exp\left(\beta^{-1} E(w) + (\beta^{-1} - 1) H(w)\right) \tag{13} \\ &= r(w) \exp\left(H(w) - \beta^{-1} KL\left(q_w(Z) \| p(Z)\right)\right) \\ &= r(w) \exp^{1/\beta}\left(\mathbb{E}_{q_w(Z)}[\log p(Z) + (\beta - 1) \log q_w(Z)]\right), \end{aligned}$$

where $H(w) = -\mathbb{E}_{q_w(Z)}[\log q_w(Z)]$ is the entropy of $q_w$ and $E(w) = \mathbb{E}_{q_w(z)}[\log p(Z)]$.

Here, $r(w)$ plays a role similar to a base density in an exponential family. If one simply used $r(w) = 1$, then $p(w|z)$ may not be well-defined. Further, without a base density, the marginal $q(z)$ would depend on the particular parameterization of $w$. To see this, take single parameter $w$, and imagine re-parameterizing so the negative reals are "squashed" by a factor of two. The base density must increase by a factor of two for the negative reals to compensate in order to leave the marginal $q(z)$ unchanged under the reparameterization.

## 2.4. Latent variables in variational inference

Adjoining $p(w|z)$ to $p(z)$ to define $D_1$ above is strongly related to *auxiliary random variables* (Agakov & Barber, 2004) explored in variational inference to increase the representative power of $q$, e.g. by including latent stochastic transition operators (Salimans et al., 2015), random Gaussian process mappings (Tran et al., 2016), or hierarchical variables (Ranganath et al., 2016). Imagine doing variational inference with $q_\theta(z, w)$, where now $w$ is auxiliary and the goal is to set $\theta$ so that $q_\theta(z) \approx p(z)$. Since $w$ must be integrated out, the entropy of $z$ under $q$ is typically intractable, but can be bounded by augmenting $p$ with some distribution $p(w|z)$ and then optimizing the joint KL-divergence over $z$ and $w$, similarly to Lemma 2. Note two differences. First, here the distribution over $w$ is mathematically derived to minimize the bound, while previous work numerically optimizes $\theta$ at run-time. Second, previous work also optimizes parameters of the distribution $p(w|z)$ to further improve the bound, while here it is left fixed (for each $\beta$) for simplicity (Sec. 5). Future work might explore optimizing $p(w|z)$ at run-time in the current paper's framework.

## 3. Bayesian Inference

This section considers algorithms to sample from the distribution defined by Eq. 13. Probabilistic inference can be used in various settings, but for concreteness the rest of this paper will focus on Bayesian inference. To fix notation, suppose a set of $N$ inputs $x^i$ with corresponding outputs $y^i$. (In a generative setting, $x^i$ would be empty.) Then, if $z$ is a vector of parameters, the posterior distribution over $z$ is, up to a constant

$$\log p(z) = \log p_0(z) + \sum_{i=1}^{N} \log p(y^i | x^i, z) + C, \tag{14}$$

where $p_0(z)$ is the prior, and $p(y^i | x^i, z)$ is the likelihood for the $i$-th datum. The goal of probabilistic inference is to be able to evaluate expectations with respect to $p(z)$.

### 3.1. Langevin Dynamics

The goal of MCMC methods is to obtain samples from the target distribution $p(z)$. Langevin dynamics sample by an extremely simple process of repeating gradient steps of $\log(z)$ with injected Gaussian noise. Specifically, the iterate is

$$z \leftarrow z + \frac{\epsilon}{2} \nabla \log p(z) + \sqrt{\epsilon} \eta, \tag{15}$$

where $\eta$ is sampled from a standard Multivariate Gaussian distribution and $\epsilon$ is a step-size that may decay over time. If Langevin dynamics are used as a proposal for a Metropolis-Hastings sampler, it can be shown that correct acceptance ratio is

$$\begin{aligned} \exp\big(s(z') - s(z) + \frac{\epsilon}{8} \|\nabla s(z)\|^2 - \frac{\epsilon}{8} \|\nabla s(z')\|^2 \\ + \frac{1}{2}(z - z') \cdot (\nabla s(z) + \nabla s(z'))\big), \end{aligned} \tag{16}$$

where $s(z) = \log p(z)$ up to a constant factor, and $z'$ is the proposed point from Eq. 15. It is easy to see that as $\epsilon$ becomes small, this acceptance ratio will go to one, and so one can disregard the acceptance step with some bias in the results determined by the step size.

Langevin dynamics explore the space of $z$ through a random walk, and thus are likely to be slower than alternatives such as Hamiltonian Monte Carlo when used as an exact method (Neal, 2010, Section 5.2). The main practical advantage of Langevin dynamics comes from the case where the number of data $N$ is large. In that case, one can use a minibatch of $M$ elements and approximate $\nabla \log p(z)$ as

$$\nabla \log p(z) \simeq \nabla \log p_0(z) + \frac{N}{M} \sum_{i \in \text{minibatch}} \nabla \log p(y^i | x^i, z). \tag{17}$$

Thus, Stochastic Gradient Langevin Dynamics, avoid a full pass through the dataset in each iteration, and so can scale to large datasets. Though Eq. 17 is an unbiased estimate of the gradient of $\log p(z)$ this still adds a bias to the stationary distribution of the the chain. (See (Mandt et al., 2016) for an analysis.) In the limit of a small step-size, the variance of the noise due to stochastic estimation of the gradient of $\log p(z)$ will be of order $\epsilon^2$ while the variance of the injected noise is of order $\epsilon$, meaning the latter dominates (Welling & Teh, 2011; Teh et al., 2016).

## 3.2. Stochastic Gradient Variational Inference

The goal of variational inference is to maximize

$$\mathcal{L}(w) = \mathbb{E}_{q_w(Z)}[\log p(Z) - \log q_w(Z)], \tag{18}$$

equivalent to minimizing the $KL$ divergence between $q_w(z)$ and $p(z)$. If it were possible to exactly compute $\mathcal{L}$, this could be maximized (to a local optima) by a simple gradient iteration like

$$w \leftarrow w + \frac{\epsilon}{2} \nabla_w \mathcal{L}(w). \tag{19}$$

While in some cases with specific $p$ and $q$, one can derive exact updates of $\mathcal{L}$ (Ghahramani & Beal, 2000) in general one cannot exactly evaluate the expectation over $Z$ in $\mathcal{L}$. One line of approach to this (Ranganath et al., 2014; Salimans & Knowles, 2014) is to write the gradient as $\nabla \mathcal{L} = \mathbb{E}_{q_w(Z)}[(\log q_w(Z) - \log p(Z)) \nabla \log q_w(Z)]$, and estimate this by drawing samples from $q_w(z)$. This experimentally seems to result in gradients with large variance, but this can be reduced by two strategies. Firstly, one can use control variates, based on either Taylor expansion (Paisley et al., 2012) or the fact that the expected value of $\nabla_w \log q_w(Z)$ is zero. Secondly, since $\log p(z)$ is often a sum of terms defined on subsets of variables, one can Rao-Blackwellize by integrating out other variables. This approach only needs to be able to compute $\log p(z)$– no other access, even to the gradient of $p$, is needed.

### 3.2.1. REPARAMETERIZATION TRICK

Another line of approach to variational inference is based on the "reparameterization trick" (Kingma & Welling,

2014; Rezende et al., 2014; Titsias & Lázaro-Gredilla, 2014), known in the stochastic approximation literature as a "pathwise" derivative (Kushner & Yin, 2003, Sec. 2.5.1). Suppose that there is a deterministic mapping $z_{r,w}$ from parameters $w$ and random numbers $r$ (from some fixed standard distribution) such that $z_{r,w} \sim q_w(z)$. Then, we could equivalently write

$$\mathcal{L}(w) = \mathbb{E}_R[\log p(z_{R,w})] + H(w). \tag{20}$$

where $H(w) = -\mathbb{E}_{q_w(Z)} \log q_w(Z)$ denotes the entropy. The advantage of Eq. 20 this is that the expectation is not a function of $w$, and so if computing the gradient of $\mathcal{L}$, the gradient moves inside the expectation. Then, provided $z_{r,w}$ is differentiable with respect to $w$, an unbiased estimate of the gradient of $\mathcal{L}$ is available as

$$\nabla_w \mathcal{L}(w) \simeq \nabla_w \log p(z_{r,w}) + \nabla_w H(w). \tag{21}$$

If $H(w)$ cannot be integrated in closed-form an alternative estimator based on

$$\mathcal{L}(w) = \mathbb{E}_R[\log p(z_{R,w}) - \log q_w(z_{R,w})] \tag{22}$$

is possible, and in some circumstances this can even have lower variance (Roeder et al., 2016).

Computing the gradient of $\log p(z_{r,w})$ with respect to $w$ amounts to computing the derivative of $\log p(z)$ with respect to $z$ multiplied by the Jacobian $dz_{r,w}^T / dw$. For simple distributions like Gaussians $z_{r,w}$ this is easy to do analytically (Section 5). Alternatively, this can all be done efficiently by automatic differentiation, (Kucukelbir et al., 2017) the approach used here.

As with Langevin dynamics, with a large dataset, $\log p(z)$ can be approximated by taking a sum over a minibatch. The most obious approach to this would be to choose a single vector $r$ and use it throughout the minibatch. However, variance can often be reduced by the "local reparameterization trick" (Kingma et al., 2015) in which a different random vector $r_i$ is drawn for each datum in the minibatch. Intuitively, the motivation is that if $i$ and $j$ are two data in the minibatch then $\log p(y^i | x^i, z_{r,w})$ should be less correlated with $\log p(y^j | x^j, z_{r',w})$ when a different random vector $r'$ is used for the second datum (consider the limiting case where all data are identical). Then, one can write the approximation as

$$\nabla_w \mathcal{L}(w) \simeq \frac{1}{M} \sum_{i \in \text{minibatch}} \Big( \nabla_w \log p_0(z_{r_i,w})$$
$$+ N \nabla_w \log p(y^i | x^i, z_{r_i,w}) - \nabla_w \log q_w(z_{r_i,w}) \Big). \tag{23}$$

Note that the approximation on the right-hand side depends both on the choice of the minibatch and on the random vectors $r_i$, one chosen per element of the minibatch.

## 4. Hybrid Dynamics

An algorithm that interpolates between the methods in the previous section results from applying Langevin dynamics to the distribution over $w$ defined by Thm. 3 If the adjoining distribution $p(w|z)$ is chosen as in Sec. 2.3 with some $r_\beta(w)$ that depends on $\beta$, then define

$$\mathcal{L}(w) = \beta \log r_\beta(w) + \mathbb{E}_{q_w(Z)}[\log p(z) + (\beta - 1) \log q(z|w)], \quad (24)$$

which is $\beta$ times the form of $s(w)$ derived in 5, with the constant term of $\log r_z$ dropped. Applying Langevin Dynamics[2] to this results in the iteration

$$w \leftarrow w + \frac{\epsilon}{2} \nabla \mathcal{L}(w) + \sqrt{\epsilon\beta}\eta, \quad (25)$$

where again $\epsilon$ is a step-size that may decrease over time and $\eta$ is sampled from a standard Gaussian distribution.

This paper uses a closed-form for the entropy $H(w) = -\mathbb{E}_{q_w(Z)} \log q_w(Z)$, and so uses the gradient estimator

$$\nabla \mathcal{L}(w) \simeq \beta \nabla \log r_\beta(w) + (1 - \beta) \nabla H(w) + \mathbb{E}_{q_w(Z)}[\log p(z) + (\beta - 1) \log q(z|w)]. \quad (26)$$

## 5. Specifics For Gaussians

The following experiments use the simplest common variational distribution for $q_w(Z)$, namely a fully-factorized Gaussian distribution. To make $w$ unconstrained, let $w = (\mu, \nu)$ where $\mu$ is a vector of mean components and the standard deviation of the $i$-th dimension is $\sigma_i = 10^{\nu_i}$. To sample from this distribution given a vector $r$, one can simply take $z_{r,w} = \mu + r \odot \sigma$ where $\odot$ is element-wise product. The entropy of $q_w(Z)$ is, up to constant factors $H(w) = \sum_i \nu_i \ln 10$.

It remains to set the base density. These experiments used an improper prior $r_\beta(w) \propto \prod_i \mathcal{N}(\nu_i|u_\beta, 1)$ that is uniform over $\mu$ with a Gaussian prior on $\nu$ with a fixed variance of 1 and a mean $u_\beta$. Since this is uniform over $\mu$, $r_z$ is a constant. The value $u_\beta$ was calculated to numerically optimize the divergence bound $D_\beta^*$ for a one-dimensional standard Gaussian $p(z)$. Since $D_\beta^* = -\beta A$ at the solution (Theorem 3), for any given $\beta$ and $u_\beta$, $D_\beta^*$ can be evaluated by symbolically integrating out $\mu$ and then numerically integrating over $\nu$. The values used in these experiments are below, with linear interpolation used for any other $\beta$.

| $\beta$ | 0 | 0.1 | 0.2 | 0.3 | 0.4 | 0.5 | 0.6 | 0.7 | 0.8 | 0.9 | 1.0 |
|---|---|---|---|---|---|---|---|---|---|---|---|
| $u_\beta$ | -.33 | -.472 | -.631 | -.792 | -.953 | -1.11 | -1.29 | -1.49 | -1.74 | -2.10 | -10 |

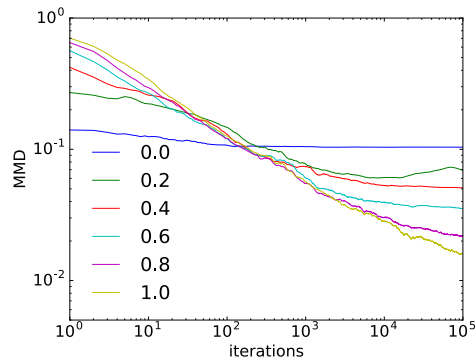

*Figure 2.* Performance of the hybrid algorithm with various values of $\beta$ on the toy mixture of three Gaussians from Fig. 1.

Theoretically, it's easy to show that as $\beta \to 1$, $D_\beta^*$ is minimized (with a minimum of zero) by $u_\beta \to -\infty$, meaning that after one iteration, $\nu = -\infty$, and henceforth, $z_{r,w} = \mu$. Thus, Eq. 25 is equivalent to Langevin dynamics as in Eq. 15. However, using $u_\beta = -10$, $r(w)$ (a prior centered on a standard deviation of $\sigma = 10^{-10}$) is practically equivalent and is more useful to guide linear interpolation. Meanwhile, when $\beta \to 0$, the algorithm is independent of $r(w)$, though the optimal value of $u_\beta \approx .33$ is correct for small $\beta$.

The above choice of $r_\beta(w)$ is quite arbitrary. It is possible that other choices could be better, though performance was surprisingly insensitive in practice. With large datasets $\log p(z)$ dominates $\log r(w)$ just as likelihoods dominate Bayesian priors. Calling on inspiration from recent work in variational inference, it might also be useful to optimize $p(w|z)$ at runtime (Section 2.4).

## 6. Experiments

There does not appear to be a single standard performance measure to evaluate approximate inference algorithms. For Bayesian inference, test-set accuracy or likelihood are sometimes used, but these measures have high variance due to the dataset, and conflate evaluation of the inference method with evaluation of the model it is running on. (An inference method with poor coverage of $p(z)$ can have higher accuracy than a perfect sampler due to model mis-specification or dataset peculiarities.)

For a more fine-grained measure of inference performance, this paper uses the Maximum Mean Discrepancy (MMD) measure (Gretton et al., 2006; 2012). MMD is essentially the empirical difference of the means of two samples in some feature space. We first draw a sample from $p(z)$ by exhaustively running a traditional MCMC algorithm (Stan (Team, 2016), based on a variant of Hamil-tonian Monte Carlo (Hoffman & Gelman, 2014)) for a large number of iterations, and compare this to the results of approximate inference. Since we have large samples, and want to compute the online performance at all iterations, a finite-dimensional feature space must be used to avoid the quadratic complexity of computing MMD using an arbitrary kernel. For two-dimensional problems, it was beneficial to use random Fourier features (Rahimi & Recht, 2007) to approximate the radial-basis function kernel $k(x, y) = \exp(-\gamma \|x - y\|^2)$. For higher dimensions, the original feature space was sufficiently discriminative.

---

[2]For some step-size $\epsilon'$, the Langevin dynamics would immediately be $w \leftarrow w + (\epsilon'/2)\nabla s(w) + \sqrt{\epsilon'}\eta$. If we define $\epsilon = \beta\epsilon'$, then this results in the given form since $\epsilon'\nabla s(w) = (\epsilon/\beta)\nabla s(w) = \epsilon\nabla \mathcal{L}(w)$.

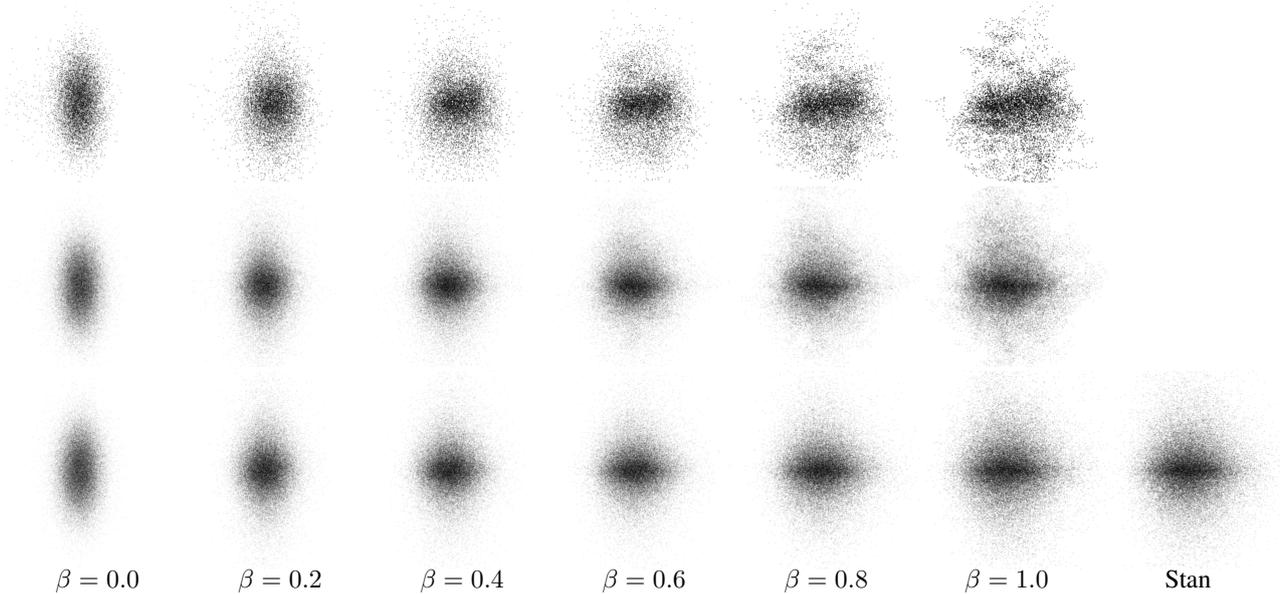

*Figure 3.* Samples on `ionosphere` after $10^4$ (top row), $10^5$ (middle row) or $10^6$ (bottom row) iterations. One sample $z$ is drawn from $q_w(z)$ at each iteration and projected onto the first two principal components (computed on Stan samples). Each plot uses the step-size $\epsilon$ with the lowest MMD at that time horizon. The same sequence of random numbers is for all inference methods. (More in appendix.)

Approximate inference gives a sequence of vectors $w_t$. For each time-step, a set of 100 samples $A_t$ are drawn from $q(z|w_t)$. Then, the MMD between $A_1 \cup A_2 \cup \cdots A_t$ and the sample from exhaustive MCMC can be computed for all $t$ in linear time. A disadvantage of this approach is the need for exhaustive MCMC for comparison, which by definition eludes the large-scale cases that motivate Langevin dynamics and stochastic variational inference. However this gives a more fine-grained view of performance.

### 6.1. Toy Distributions

For a first demonstration, the algorithm was applied to a set of toy 2-dimensional distributions, with various values of $\beta$. A few results are shown in Fig. 1, with more in the appendix. While the algorithm displays the expected trade-offs between speed and accuracy for different $\beta$ (Fig. 2), the results fairly uninteresting as all curves cross near the same time/MMD point where $\beta = 0$ (variational inference) and $\beta = 1$ (MCMC) meet. So, for any given amount of time, either variational inference or MCMC are near-optimal, and so intermediate $\beta$ do not much expand the "Pareto frontier" on these problems, although one might prefer an intermediate $\beta$ since the crossover horizon is not known in advance.

### 6.2. Bayesian Logistic Regression

Next, the algorithm was applied to binary logistic regression with several standard datasets, using a minibatch size of 25, a different random vector $r$ for each element in the minibatch, and a standard multivariate Laplace distribution as a prior. Simply picking a single value or schedule for the step $\epsilon$ is unsatisfactory, as the best schedule for a given problem, length of time, and value of $\beta$ varies. To give a fair comparison, inference was run for with a range of constant $\epsilon$ varying by factors of two from $2^3/N$ to $2^{-2}/N$. This spans the range from large enough to often diverge to below optimal even after $10^6$ iterations. Then, for each time, the best value of $\epsilon$ was retrospectively chosen (with performance averaged over repetitions before this choice). It is likely that decaying step-size schedules would perform

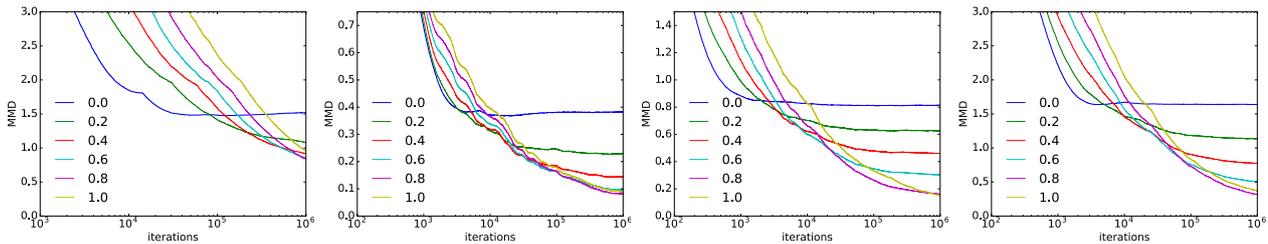

*Figure 4.* Iterations vs MMD for inference with various values of $\beta$, averaged over five runs, each using the same random number sequence for all $\beta$. For each $\beta$, iteration and dataset, the graphs show the MMD for the best step-size $\epsilon$ (averaged over the 100 runs before computing the best). Since a finite range of $\epsilon$ varying by factors of two is used, this can result in visible "kinks" at iterations where a different (smaller) $\epsilon$ exceeds the old one. From left to right: `a1a`, `australian`, `ionosphere`, and `sonar`.

better, but this was not pursued since it would reduce transparency of the results.

The mean errors after 100 repetitions are shown in Fig. 4, while samples are compared to the ground truth for ionosphere in Fig. 3. This confirms the basic ideas of the algorithm, namely that it smoothly interpolates between the two previous algorithms without introducing any new behavior. The main experimental finding is that there is usually a large range of time horizons for which an intermediate setting of $\beta$ performs better than either of the previous existing algorithms. This suggests that, say, someone who has a time budget slightly larger than that needed for variational inference could benefit from running the algorithm with a small value of $\beta$ for a slightly larger time. This is in contrast to the toy two-dimensional examples where for almost all time horizons either $\beta = 0$ or $\beta = 1$ was optimal.

## 7. Discussion

This paper has two contributions. First, a distribution over variational parameters is derived to minimize a bound on the marginal divergence to a target distribution. Secondly, this is used to derive an algorithm that interpolates between Langevin dynamics and stochastic variational inference. The Bayesian logistic regression experiments show that the intermediate values of this algorithm are useful in the sense that there are several orders of magnitude of time horizons where an intermediate algorithm performs better than either Langevin dynamics or variational inference.

Many improvements to stochastic Langevin dynamics and variational inference have been proposed beyond the simple algorithms used here. For Langevin dynamics, Welling & Teh (2011) and Li et al. (2016) suggest adding a preconditioner, Patterson & Teh (2013) propose Riemannian Langevin Dynamics, and Dubey et al. (2016) propose to make use of the finite sum structure in a Bayesian posterior $p(z)$ by incorporating techniques for incremental optimization. For stochastic variational inference, Hoffman et al.

(2013) propose to use the natural gradient, Mandt & Blei (2014) propose using smoothed gradients, and Sakaya & Klami (2016) propose a strategy of re-using previous gradient computations. It would be interesting to consider using these ideas with the hybrid algorithm, which could give insight into the relationship between the two different methods.

There is other work based on combining the advantages of variational inference and MCMC. Stochastic Gradient Fisher Scoring (Ahn et al., 2012) interpolates between Langevin dynamics and a Gaussian approximation of the target. However, this approximation is based on the Bayesian central limit theorem and so will not be the same in general as the optimal variational distribution, even when a Gaussian is used as the variational family. Another line of research is that of interpreting the path of a MCMC algorithm as a variational distribution, and then fitting parameters to tighten a variational bound (Salimans et al., 2015; Rezende & Mohammed, 2015). While motivationally quite similar, the contribution of this paper is orthogonal, in that one could conceivably use these same ideas to define the variational family $q_w(z)$ used in this paper. This would result in a somewhat unusual method that runs MCMC over the parameters of variational family of distributions that are themselves defined in terms of a (different) MCMC algorithm.

## 8. Appendix

This appendix contains additional plots and proofs of the results from Section 2.

**Lemma 6.** *The divergence from $q(z)$ to $p(z)$ is*

$$KL\left(q(Z)\|p(Z)\right) = \underbrace{KL\left(q(Z|W)\|p(Z)\right)}_{D_0} - I_q[W, Z],$$
(27)

*where $D_0 = \mathbb{E}_{q(W,Z)} \log\left(q(Z|W)/p(Z)\right)$ is conditional divergence and $I_q$ denotes mutual information under $q$.*

*Proof.* Define the joint distribution $p(w, z) = q(w)p(z)$. Then, the chain-rule of KL-divergence (Cover & Thomas, 2006, Thm. 2.5.3) states that

$$KL\left(q(Z,W)\|p(Z,W)\right) = KL\left(q(W|Z)\|p(W|Z)\right) + KL\left(q(Z)\|p(Z)\right). \quad (28)$$

The left-hand side simplifies into $D_0$, and the first term on the right-hand side simplifies into $I_q[W, Z]$. □

**Theorem 7.** *For fixed values of $\beta$ and $p(w|z)$, the distribution $q(w)$ that minimizes $D_\beta$ is*

$$q^*(w) = \exp\left(s(w) - A\right)$$
$$A = \log \int_w \exp s(w)$$
$$s(w) = \log p(w) - KL\left(q(Z|w)\|p(Z|w)\right)$$
$$- \left(\beta^{-1} - 1\right) KL\left(q(Z|w)\|p(Z)\right).$$

*Moreover, at $q^*$, the objective value is $D_\beta^* = -\beta A$.*

*Proof.* First, consider derivatives of $D_0$ and $D_1$ with respect to $q(w)$. The first can immediately be seen to be

$$\frac{dD_0}{dq(w)} = KL\left(q(Z|w)\|p(Z)\right).$$

For the second, we can derive

$$\frac{dD_1}{dq(w)} = \frac{d}{dq(w)} \int_{w,z} q(w, z) \log \frac{q(z|w)}{p(w, z)}$$
$$+ \frac{d}{dq(w)} \int_{w,z} q(w) \log q(w)$$
$$= \int_z q(z|w) \log \frac{q(z|w)}{p(w, z)} + \log q(w) + 1$$
$$= KL\left(q(Z|w)\|p(Z|w)\right) - \log p(w) + \log q(w) + 1.$$

If we create a Lagrangian for $D_\beta$ with a Lagrange multiplier $\lambda$ to enforce normalization of $q(w)$, we know that at

the optimal $q(w)$ its gradient will be zero. Using the above derivatives, we therefore have that

$$0 = (1 - \beta)KL\left(q(Z|w)\|p(Z)\right) + \beta KL\left(q(Z|w)\|p(Z|w)\right)$$
$$- \beta \log p(w) + \beta \log q(w) + \lambda,$$

Which solved for $q(w)$, this gives

$$q(w) \propto \exp\left(- (1 - \beta^{-1})KL\left(q(Z|w)\|p(Z)\right)\right.$$
$$\left. - KL\left(q(Z|w)\|p(Z|w)\right) + \log p(w)\right),$$

which establishes the given form for $s(w)$ and $A$.

Now, to establish the value of $D_\beta$ at the solution, expand the negative entropy of $q(w)$ to get

$$\beta \int_w q(w) \log q(w)$$
$$= \beta \int_w q(w) \Big(- (1 - \beta^{-1})KL\left(q(Z|w)\|p(Z)\right)$$
$$- KL\left(q(Z|w)\|p(Z|w)\right) + \log p(w)\Big) - \beta A. \quad (29)$$

Now, taking the left-hand side and terms in the bottom line, we can recognize that

$$\int_w q(w) \left(\log \frac{p(w)}{q(w)} - KL\left(q(Z|w)\|p(Z|w)\right)\right) = -D_1.$$

Further, if we take the terms from the middle line, we have that

$$-\beta \int_w q(w)(1 - \beta^{-1})KL\left(q(Z|w)\|p(Z)\right) = (\beta - 1)D_0.$$

Thus, we can re-write Eq. 29 as $-\beta A = (1 - \beta)D_0 + \beta D_1$, establishing the value of $D_\beta^*$. □

*Remark 8.* In the limit where $\beta \to 0$ the divergence bound becomes

$$\lim_{\beta \to 0} D_\beta^* = \inf_w KL\left(q(Z|w)\|p(Z)\right).$$

*Proof.* Use the representation that $\lim_{\beta \to 0} D_\beta^* = \lim_{\beta \to 0} -\beta A$ is equal to

$$\lim_{\beta \to 0} -\beta \log \int_w \exp\Big(\log p(w) - KL\left(q(Z|w)\|p(Z|w)\right)$$
$$- \left(\beta^{-1} - 1\right) KL\left(q(Z|w)\|p(Z)\right)\Big)$$
$$= \lim_{\beta \to 0} -\beta \log \int_w \exp\Big(-\beta^{-1}KL\left(q(Z|w)\|p(Z)\right)\Big).$$

The form for $D_\beta^*$ follows from the fact that $\lim_{\beta \to 0} \beta \log \int_w \exp(\beta^{-1}f(w)) = \sup_w f(w)$. □

**Lemma 9.** *If $p(w|z) = r(w)q(z|w)/r_z$ and $r_z$ is a constant, then the solution in Thm. 3 holds with*

$$
\begin{aligned}
s(w) = {} & \log r(w) - \log r_z \\
& + \mathbb{E}_{q_w(Z)}[\beta^{-1} \log p(z) + (1 - \beta^{-1}) \log q(z|w)].
\end{aligned}
$$

*Proof.* First, without using the particular form for $p(w|z)$, we can write $s(w)$ as

$$
\begin{aligned}
\log p(w) - \int_z q(z|w) \log \frac{q(z|w)}{p(z|w)} \\
- \left(\beta^{-1} - 1\right) \int_z q(z|w) \log \frac{q(z|w)}{p(z)}
\end{aligned}
$$

Cancelling terms involving $q(z|w)$ in the numerators, this is

$$
\begin{aligned}
\log p(w) - \int_z q(z|w) \log \frac{p(z)}{p(z|w)} \\
- \beta^{-1} \int_z q(z|w) \log \frac{q(z|w)}{p(z)}
\end{aligned}
$$

The $\log p(w)$ can be absorbed into the first term to give, after some cancellation that

$$
s(w) = \int_z q(z|w) \log p(w|z) - \beta^{-1} KL\left(q(Z|w) \| p(Z)\right).
$$

Now, using the assumed form for $p(w|z)$, we can immediately write that $s(w)$ is

$$
\int_z q(z|w) \log \frac{r(w)q(z|w)}{r_z} - \beta^{-1} \int_z q(z|w) \log \frac{q(z|w)}{p(z)},
$$

equivalent to the form stated. ∎

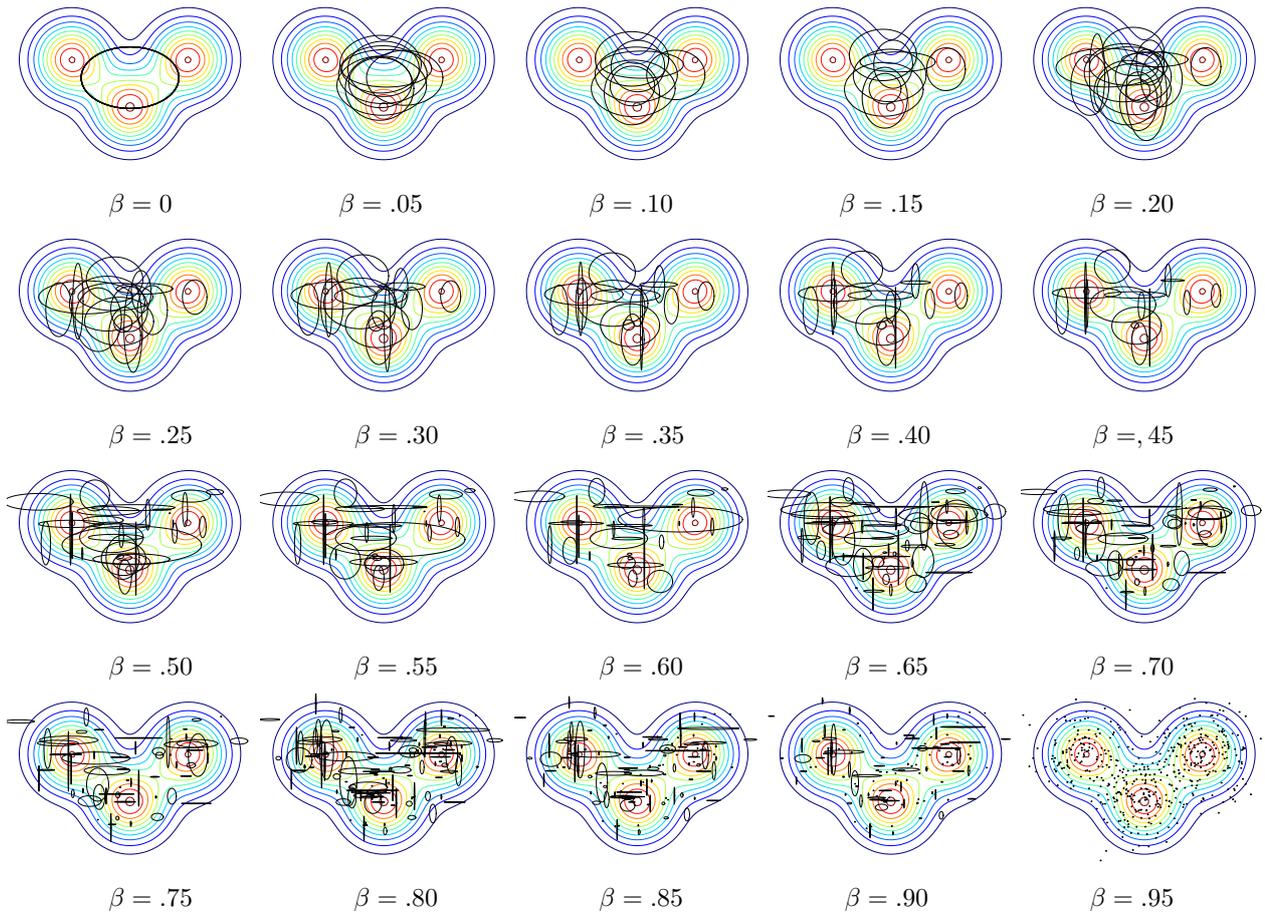

*Figure 5.* Examples sampling from a two-dimensional mixture of three gaussians after running inference for $5 \times 10^5$ iterations. The sampled weights $w$ are pictured as ellipsoids at one standard deviation. Colored contours show the density $p(z)$. To avoid visual clutter, a smaller number (equally spaced) of samples are shown for smaller $\beta$.

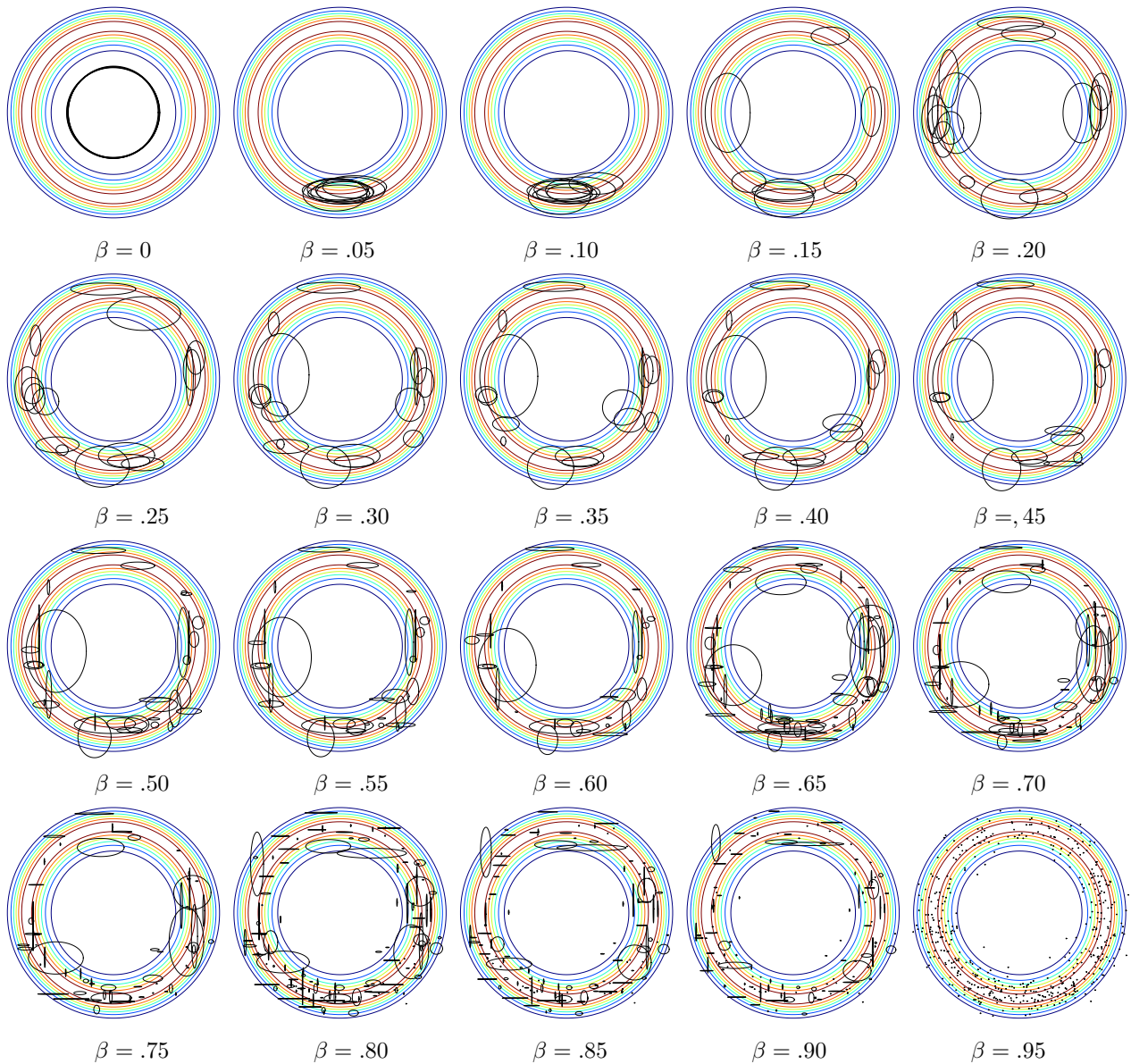

*Figure 6.* Examples sampling from a two-dimensional "donut" distribution after running inference for $5 \times 10^5$ iterations. The sampled weights $w$ are pictured as ellipsoids at one standard deviation. Colored contours show the density $p(z)$. To avoid visual clutter, a smaller number (equally spaced) of samples are shown for smaller $\beta$.

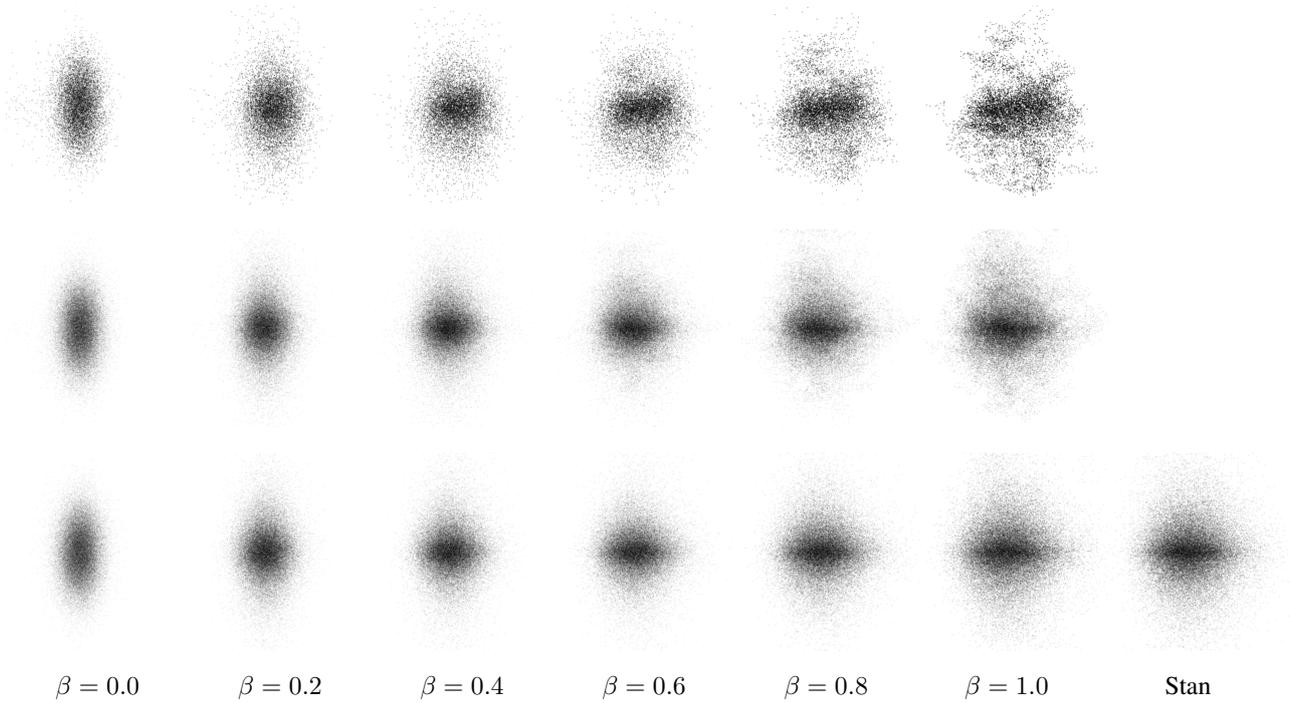

*Figure 7.* Inference for various values of $\beta$ on `ionosphere` after $10^4$ (top row) $10^5$ (middle row) or $10^6$ (bottom row) iterations. After each iteration, one sample is drawn from $q_w(Z)$, and plots show the first two principal components (computed on samples from Stan). Each plot show samples resulting from the (constant) step-size $\epsilon$ that resulted in the minimum MMD for that $\beta$ and number of iterations. The same sequence of random numbers is for all inference methods. (More results are in the appendix.)

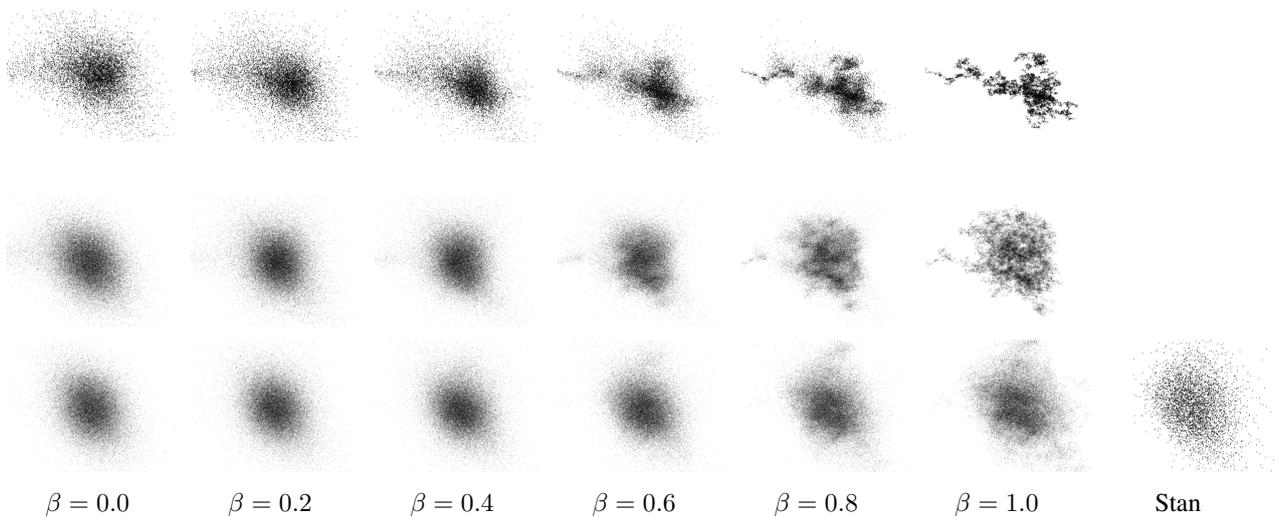

*Figure 8.* Inference for various values of $\beta$ on `ala` after $10^4$ (top row) $10^5$ (middle row) or $10^6$ (bottom row) iterations. In some of these plots, a "tail" is visible, reflecting the path into the high-density region from where $w = 0$ where inference was initialized.

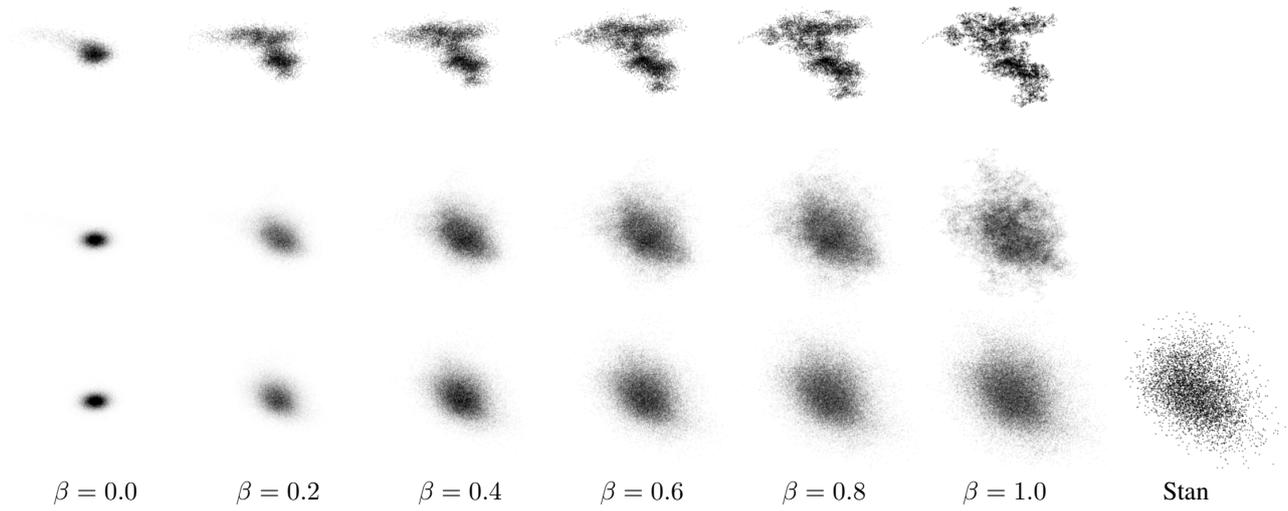

*Figure 9.* Inference for various values of $\beta$ on `australian` after $10^4$ (top row) $10^5$ (middle row) or $10^6$ (bottom row) iterations.

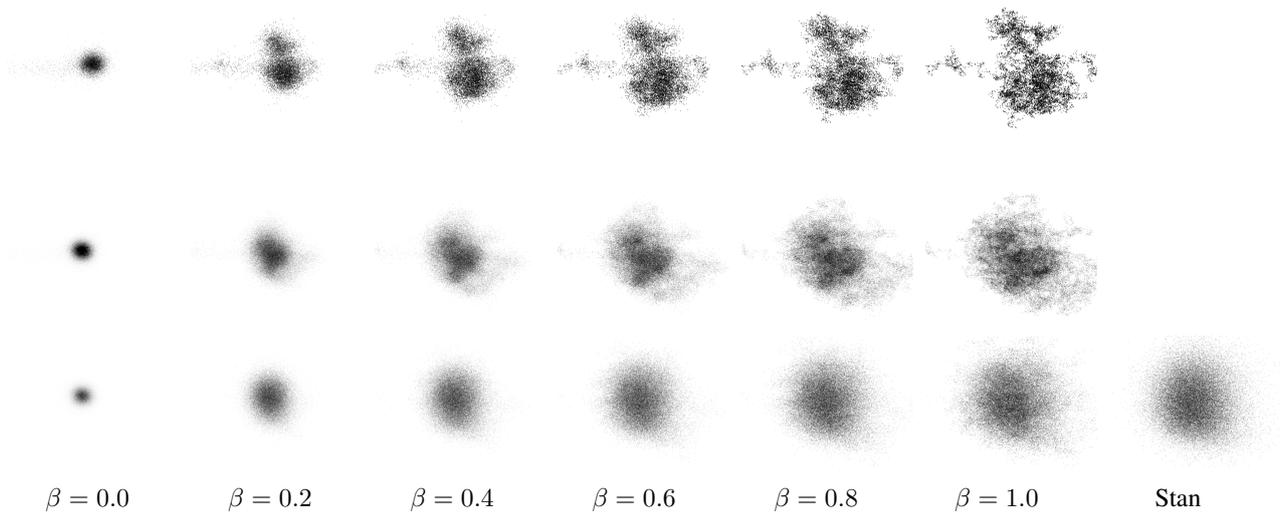

*Figure 10.* Inference for various values of $\beta$ on `sonar` after $10^4$ (top row) $10^5$ (middle row) or $10^6$ (bottom row) iterations.